\documentclass[11pt,letterpaper]{article}
\usepackage[letterpaper]{geometry}
\usepackage{acl2012}
\usepackage{times}
\usepackage{latexsym}
\usepackage{amsmath}
\usepackage{multirow}
\usepackage{url}
\makeatletter
\newcommand{\@BIBLABEL}{\@emptybiblabel}
\newcommand{\@emptybiblabel}[1]{}
\makeatother
\usepackage[hidelinks]{hyperref}

\usepackage{graphicx}
\usepackage{fancyvrb}
\usepackage{framed}
\usepackage[inline]{enumitem}

\graphicspath{ {images/} }

\title{\#HashtagWars: Learning a Sense of Humor}

\author{Peter Potash, Alexey Romanov, Anna Rumshisky \\
  University of Massachusetts Lowell \\
  Department of Computer Science \\
  {\tt \{ppotash,aromanov,arum\}@cs.uml.edu} }

\date{}

\begin{document}
\maketitle

\begin{abstract}
In this work, we present a new dataset for computational humor, specifically
comparative humor ranking, which attempts to eschew the ubiquitous binary approach to humor detection. The dataset consists of tweets that are humorous responses to a given hashtag. We describe the motivation for this new dataset, as well as the collection process, which includes a description of our semi-automated system for data collection. We also present initial experiments for this dataset using both unsupervised and supervised approaches. 
%
%
%
%
Our best supervised system achieved 63.7\% accuracy, suggesting that this task is much more difficult than comparable humor detection tasks. 
Initial experiments indicate that a character-level model is more suitable for this task than a token-level model, likely due to a large amount of puns that can be captured by a character-level model.

\end{abstract}

\section{Introduction}


Most work on humor detection approaches the problem as binary classification: humor or not humor. While this is a reasonable initial step, in practice humor is subjective, so we believe it is interesting to evaluate different degrees of humor, particularly as it relates to a given person's sense of humor. To further such research, we propose a dataset based on humorous responses submitted to a Comedy Central TV show, allowing for computational approaches to comparative humor ranking.


Debuting in Fall 2013, the Comedy Central show @midnight\footnote{http://www.cc.com/shows/-midnight} is a late-night ``game-show'' that presents a modern outlook on current events by focusing on content from social media. The show's contestants (generally professional comedians or actors) are awarded points based on how funny their answers are. The segment of the show that best illustrates this attitude is the Hashtag Wars (HW). Every episode the show's host proposes a topic in the form of a hashtag, and the show's contestants must provide tweets that would have this hashtag. Viewers are encouraged to tweet their own responses. From the viewers' tweets, 
we are able to apply labels that determine how relatively humorous the show finds a given tweet.

Because of the contest's format, it provides an adequate method for addressing the selection bias \cite{heckman1979sample} often present in machine learning techniques \cite{zadrozny2004learning}. Since each tweet is intended for the same hashtag, each tweet is effectively drawn from the same sample distribution. Consequently, tweets are seen not as humor/non-humor, but rather varying degrees of wit and cleverness. Moreover, given the subjective nature of humor, labels in the dataset are only ``gold'' with respect to the show's sense of humor. This concept becomes more grounded when considering the use of supervised systems for the dataset.

The goal of the dataset is to learn to characterize the sense of humor represented in this show.  Given a set of hashtags, the goal is to predict which tweets the show will find funnier within each hashtag. The degree of humor in a given tweet is determined by the labels provided by the show.  We evaluate potential predictive models based on a pairwise comparison task in an initial effort to leverage the HW dataset. The pairwise comparison task will be to select the funnier tweet, and the pairs will be derived from the labels assigned by the show to individual tweets. 
Initial experiments on the HW dataset will involve both unsupervised and supervised approaches.

There have been numerous computational approaches to humor within the last decade \cite{yanghumor,mihalcea2005making,zhang2014recognizing,radev2015humor,raz2012automatic,reyes2013multidimensional,barbieri2014automatic,shahaf2015inside,purandare2006humor,kiddon2011s}. In particular, \cite{zhang2014recognizing,raz2012automatic,reyes2013multidimensional,barbieri2014automatic} focus on recognizing humor in twitter. However, the majority of this work decomposes the notion of humor into two groups: humor and non-humor. This representation ignores the continuous nature of humor, while also not accounting for the subjectivity in perceiving humor. Humor is an essential trait of human intelligence that has not been addressed extensively in the current AI research, and we feel that shifting from the binary approach of humor detection is a good pathway towards advancing this work. 


To further motivate the need for a task that acknowledges the subjective nature of humor, we report the results of an annotation task from Shahaf et al. \shortcite{shahaf2015inside}. The authors asked annotators to look at pairs of captions from the New Yorker Caption Content\footnote{\url{http://contest.newyorker.com/}} (for more information on the dataset, see Section \ref{sec:related_work}). Unfortunately, the authors report, ``Only 35\% of the unique pairs that were ranked by at least five people achieved 80\% agreement...''. This statistic further supports the notion that sense of humor is not an objective linguistic quality. Consider the task of semantic relatedness, which is a far more subjective task than
part-of-speech tagging. Even for this task, which requires
a strong amount of individual interpretation,the average standard deviation for relatedness scores (in the range 1-5) was 0.76~\cite{marelli2014sick}, which conveys a low
disagreement. Sense of humor is a truly unique quality to
each individual, and language is more the means used to communicate that sense of humor. Therefore, data-driven
approaches for understanding humor must acknowledge
the individual nature of humor taste, and not treat it as
a universal notion such as language itself.


The broader impact of our dataset will be in the field of human-computer interaction. As evidence we highlight two systems that use humor in a human-computer dynamic. First, in \cite{wen2015omg} a computer chat agent attempts to suggest humorous memes/images in response to questions, creating an enjoyable experience for users. Dybala et al. \shortcite{dybala2013metaphor} offer a system that is better applicable to pure text. The system attempts to detect if the user is in a negative emotional state. If so, the computer offers humor in an effort to improve the user's mood. In terms of personalized interaction, it is not adequate to treat humor as binary, but rather as a continuous spectrum, seeking to understand the sense of humor unique to a given user.


\section{Related Work}\label{sec:related_work}

Mihalcea and Strapparava \shortcite{mihalcea2005making} developed a humor dataset of puns and humorous
one-liners intended for supervised learning. In order to generate negative examples for their
experimental design, the authors used news title from Reuters news, proverbs and British National
Corpus. Recently, Yang et al. \shortcite{yanghumor} used the same same dataset for experimental
purposes, taking text from AP News, New York Times, Yahoo! Answers and proverbs as their negative
examples. To further reduce the bias of their negative examples, the authors selected negative
examples with a vocabulary that is in the dictionary created from the positive examples. Also, the
authors forced the negative examples to have a similar text length compared to the positive
examples.

Zhang and Liu \shortcite{zhang2014recognizing} constructed a dataset for recognizing
humor in Twitter in two parts. First, the authors use the Twitter API with target user mentions
and hashtags to produce a set of 1,500 humorous tweets. After manual inspections, 1,267 of the
original 1,500 tweets were found to be humorous, of which 1,000 were randomly sampled as positive
examples in the final dataset. Second, the authors collect negative examples by extracting
1,500 tweets from Twitter Streaming API, manually checking for the presence of humor. Next, the
authors combine these tweets with tweets from part one that were found to actually not contain
humor. The authors argue this last step will partly assuage the selection bias of the negative
examples.


In Reyes et al. \shortcite{reyes2013multidimensional} the authors create a model to detect ironic tweets. To construct their dataset they collect tweets with the following hashtags: irony, humor, politics, and education. Therefore, a tweet is considered ironic solely because of the presence of the appropriate hashtag. Barbieri and Saggion \shortcite{barbieri2014automatic} also use this dataset or their work.
Finally, within the last year researchers have developed a dataset similar to our HW dataset based on the New Yorker Caption contest (NYCC) \cite{radev2015humor,shahaf2015inside}. While for the HW viewers submit a tweet in response to a hashtag, for the NYCC readers submit humorous captions in response to a cartoon. It is important to note this key distinction between the two datasets, because we believe that the presence of the hashtag allows for further innovative NLP methodologies aside from solely analyzing the tweets themselves. In Radev et al. \shortcite{radev2015humor}, the authors developed more than 15 unsupervised methods for ranking submissions for the NYCC. The methods can be categorized into broader categories such as originality and content-based.

Alternatively, \newcite{shahaf2015inside} approach the NYCC dataset with a supervised model, evaluating on a pairwise comparison task, upon which we base our evaluation methodology. The features to represent a given caption fall in the general areas of Unusual Language, Sentiment, and Taking Expert Advice.
For a single data point (which represents two captions), the authors concatenate the features of each individual caption, as well as encoding the difference between each caption's vector. The authors' best-performing system records a 69\% accuracy on the pairwise evaluation task. Note that for this evaluation task, random baseline is 50\%. Therefore, the incremental improvement above random guessing dictates the difficulty of predicting degrees of humor.

\section{\#HashtagWars Dataset}

\subsection{Data collection}

The following is our data collection process. First, when a new episode airs (which generally happens four nights a week, unless the show is on break) a new hashtag will be given. We wait until the following morning to use the Twitter search API\footnote{\url{https://dev.twitter.com/rest/public/search}} to collect tweets that have been posted with the new hashtag. Generally, this returns 100-200 tweets. We wait until the following day to allow for as many tweets as possible to be submitted. The day of the ensuing episode (i.e. on a Monday for a hashtag that came out for a Thursday episode), @midnight creates a Tumblr post\footnote{\url{http://atmidnightcc.tumblr.com/}} that announces the top-10 tweets from the previous episode's hashtag. If they're not already present, we add the tweets from the top-10 to our existing list of tweets for the hashtag. We also perform automated filtering to remove redundant tweets. Specifically, we see that the text of tweets (aside from hashtags and user mentions) are not the same. The need for this results from the fact that some viewers submit identical tweets. 

Using both the @midnight official Tumblr account, as well as the show's official web-site where the winning tweet is posted, we annotate each tweet with labels \verb|0|, \verb|1| and \verb|2|. Label \verb|2| designates the winning tweet. Thus, the label \verb|2| only occurs once for each hashtag. Label \verb|1| indicates that the tweet was selected as a top-10 tweet (but \textit{not} the winning tweet) and label \verb|0| is assigned for all other tweets.
It is important to note that every time we collect a tweet, we must also collect its tweet ID. A public release of the dataset must comply with Twitter's terms of use\footnote{\url{https://dev.twitter.com/overview/terms}}, which disallows the public distribution of users' tweets. The need to determine the tweet IDs for tweets that weren't found in the initial query (i.e. tweets added from the top 10) makes the data collection process slightly laborious, since the top-10 list doesn't contain the tweet text. In fact, it doesn't even contain the text itself since it's actually an image.

\subsubsection{A Semi-Automated System for Data Collection}
Because the data collection process is continuously repeated and requires a non-trivial amount of human labor, we have built a helper system that can partially automate the process of data collection. This system is organized as a web-site with a convenient user interface.

On the start page the user enters the id of the Tumblr post with the tweets in top 10.
After that, we invoke Tesseract~\footnote{\url{https://github.com/tesseract-ocr/tesseract}}, an OCR command-line utility, to recognize the textual content of the tweets' images. Using the recognized content, the system forms a web-page on which the user can simultaneously see the text of the tweets as well as the original images. 
On this page the user can query the Twitter's API to search by text or click the button "Open twitter search" to open the Twitter Search page if the API returns zero results. 
We note that the process is not fully automated because a given text query can we return redundant results, and we primarily check to make sure we add the tweet that came from the appropriate user. With the help of this system, the process of collecting the top-10 tweets (along with their tweet IDs) takes roughly 2 minutes.
Lastly, we note that the process for annotating the winning tweet (which is already included in the top-10 posted in the Tumblr list) is currently manual, because it requiries going to the @midnight website. This is another aspect of the data collection system that could potentially be automated.


\subsection{Dataset}

\label{sec:dataset}

Data collection has been in process for roughly seven months, producing a total of 9,658 tweets for 86 hashtags. The resulting data set is currently being used in a SemEval-2017 task on humor detection.

\begin{figure}[ht]
    \centering
    \includegraphics[scale=0.4]{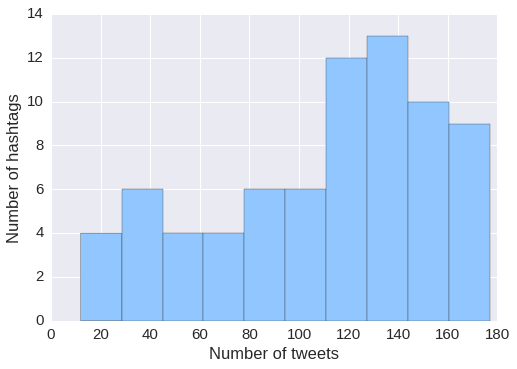}
    \caption{Distribution of the numbers of tweets per hashtag}
    \label{fig:nb_tweets_per_hashtag}
\end{figure}

%
The distribution of the number of tweets per hashtag is represented in Figure~\ref{fig:nb_tweets_per_hashtag}. For 71\% of  hashtags we have at least 90 tweets. The files of the individual hashtags are formatted so that the individual hashtag tokens are easily recoverable. Specifically, tokens are separated by the `\_' character.

Figure~\ref{fig:tweets_example} represents an example of the tweets collected for the hashtag \textit{FastFoodBooks}. Note that this hashtag requires an external knowledge about fast food and books in order to understand the humor. Furthermore, this
hashtag illustrates how prevelant puns are in the dataset,
especially related to certain hashtags. In contrast, the hashtag \textit{IfIWerePresident} (see the Figure~\ref{fig:tweets_example2}) does not require an external knowledge and the tweets are understandable without awareness about any specific concepts.

\begin{figure}[ht]
    \begin{framed}
As I Lay Dying of congestive heart failure @midnight  \#FastFoodBooks \\
Harry Potter and the Order of the Big Mac \#FastFoodBooks @midnight \\
The Girl With The Jared Tattoo \#FastFoodBooks @midnight \\
A Room With a Drive-thru @midnight \#FastFoodBooks
	\end{framed}

	\caption{An example of the items in the dataset for the hashtag \textit{FastFoodBooks} that requires external knowledge in order to understand the humor. Furthermore, the tweets for this hashtag are puns connecting book titles
    and fast food-related language}
    \label{fig:tweets_example}
\end{figure}

\begin{figure}[ht]
    \begin{framed}
\#IfIWerePresident my Cabinet would just be cats. @midnight \\
Historically, I'd oversleep and eventually get fired. @midnight \#IfIWerePresident \\
\#IfIWerePresident I'd pardon Dad so we could be together again... @midnight \\
\#IfIWerePresident my estranged children would finally know where I was @midnight \\
	\end{framed}

	\caption{An example of the items in the dataset for the hashtag \textit{IfIWerePresident} that does not require external knowledge in order to understand the humor}
    \label{fig:tweets_example2}
\end{figure}

\section{Experiments}

\subsection{Evaluation Methodology}
\label{sec:eval_methodology}
Both supervised and unsupervised approaches to this task can
be evaluated using the same consistent methodology as follows. Using the tweets submitted for each hashtag, we construct pairs of tweets in which one tweet is judged by the show to be funnier than the other.  The accuracy of prediction of the funnier tweet is then used as the evaluation measure.
The pairs used for evaluation are constructed as follows:
\begin{enumerate}
\item[(1)] The tweets that are judged to be in the top-10 funniest tweets are paired with the tweets not in the top-10.
\item[(2)] The winning tweet is paired with the other tweets in the top-10.
\end{enumerate}
If we have $n$ tweets for a given hashtag, (1) will produce $10(n-10)$ pairs, and (2) will produce 9 pairs, giving us $10n-91$ data points for a single hashtag. Constructing the pairs for evaluation in this way ensures that one of the tweets in each pair has been judged to be funnier than another. 
The first and the second tweets in a pair are shuffled based on a coin flip.

The main evaluation measure is the micro average of accuracy on the individual test hashtags. For a given hashtag, the accuracy is the number of correctly predicted pairs divided by the total number of pairs. Therefore, random guessing will produce 50\% accuracy on this task. We also include the following metrics: percentage of individual hashtags for which accuracy is above 50\%, as well as the highest/lowest accuracy across all hashtags.


\subsection{Unsupervised experiments}
\label{sec:unsuervised_exp}
The first experiments we conduct are based on unsupervised methodology. The experiments are conducted on a total of 88,494 tweet pairs from 86 different hashtags. 

\subsubsection{Metrics}
The unsupervised methodology classifies the tweet with the greater value of a  metric (feature) as the funnier tweet of the pair. Following the methodology proposed by Radev et al. \shortcite{radev2015humor}, we apply the authors' three top-performing comparison metrics, namely LexRank~\cite{erkan2004lexrank}, as well as the positive and negative sentiment of the text (tweet in our case). In order to determine the sentiment of a tweet we use the TwitterHawk system~\cite{boag2015twitterhawk}, which placed first in topic-based tweet sentiment in SemEval 2015. We used the LexRank implementation available from the sumy library\footnote{https://github.com/miso-belica/sumy}. For a given hashtag, we calculate the individual LexRank scores of the tweets.  

\subsubsection{Results}

The results of the unsupervised experiments are presented in Table~\ref{tbl:unsupervised_exp_results}. Despite the fact that the models achieved a good accuracy on several hashtags, the micro and macro averages are barely better than random guessing and even worse in the case of LexRank. We would expect that for hashtags where negative sentiment performed the best, the hashtags themselves would encapsulate some notion of negativity. In Table \ref{tbl:top_neg} we list five hashtags with the highest accuracy using the negative sentiment metric. Clearly the top-performing hashtag \textit{MakeTVShowsEvil} has a strong sense of negativity. Unfortunately, this argument is weak for the four remaining hashtags, whose accuracy doesn't vary dramatically from the top-performing hashtag. Note \newcite{shahaf2015inside} achieved an accuracy of 61\% using sentiment as an unsupervised metric for the NYCC dataset. This fact leads us to believe that the humor in the HW dataset is harder to recognize. Furthermore, their data set was much smaller and had only 754 pairs, whereas our dataset has 88k pairs.

\begin{table*}[ht]
\centering
\scalebox{1}{
\begin{tabular}{|l|l|l|l|l|l|}
\hline
\textbf{Features}	& \textbf{Acc Micro Avg} & \textbf{Acc \textgreater 0.5}	& \textbf{Max Acc}	& \textbf{Min Acc}	\\ \hline
LexRank				& 0.47					& 36\%								& 0.65				& \textbf{0.27}		\\
Negative Sentiment	& \textbf{0.51}			& \textbf{57\%}						& 0.71				& 0.22				\\
Positive Sentiment	& \textbf{0.51}			& 51\%								& \textbf{0.76}		& 0.21				\\ \hline
         
\end{tabular}
} 
\caption{The results of the unsupervised experiment. Bold indicates the best features according to the corresponding metric.}
\label{tbl:unsupervised_exp_results}
\end{table*}

\begin{table}[ht]
\centering
\scalebox{1}{
\begin{tabular}{|l|l|}
\hline
\textbf{Hashtag} & \textbf{Accuracy} \\ \hline
Make\_TV\_Shows\_Evil                 & 0.71            \\
Hungry\_Games                 & 0.70                          \\
Twitter\_In\_5\_Words                 & 0.69             \\
Sexy\_Snacks & 0.66 \\
First\_Draft\_Cartoons & 0.65 \\ \hline
\end{tabular}
} 
\caption{The hashtags with best performance with negative sentiment metric.}
\label{tbl:top_neg}
\end{table}

\subsection{Supervised Experiments}
The supervised approach truly fulfills the notion of `learning' a sense of humor, because we attempt to predict previously unseen hashtags based on a model trained on labeled tweet pairs. Unlike the unsupervised approach, a supervised system has the benefit of seeing what tweets are funnier based on the provided training data, with the hope it can generalize to hashtags not provided in the training data.

The experimental design for our supervised experiments is based on leave-one-out (LOO) evaluation. We withhold a single hashtag file for testing, and train on data generated from the remaining hashtag files. We create data points according to the methodology from Section~\ref{sec:eval_methodology}.

On average, there are 112 tweets per file. Therefore, on average we train on 87,465 data points and test on 1,029 data points. Through the course of an entire LOO experiment, we test on a total of 88,494 data points, which is the result of 86 LOO experiments.
We experiment with two different supervised methods. First, we train a feed-forward neural network (FFNN) based on hand-engineered features. Second, we experiment with a model that connects recurrent neural networks to a FFNN, with the goal of learning optimal tweet representations for our task.
In our experiments, if the first tweet is funnier, the corresponding label is \verb|1|. If the second tweet is funnier, the corresponding label is \verb|0|. We place the funnier tweet based on a coin flip, so the resulting training/test sets have roughly balanced labels.

There are three factors that lead to the creation of a data point in the supervised system: two tweets and the hashtag that prompts the two tweets. Therefore, to fully represent a data point, we believe it needs to account for the two tweets as well as the hashtag, which is a unique aspect of the HW dataset. 
In the following sections, we explain three models that we experimented with: a feed-forward neural network with hand-crafted features, a token-level recurrent neural network (RNN) model and a character-level convolutional neural network (CNN) model.

\subsubsection{Feed-Forward Neural Network Model}
As the base classifier, we used a fully connected neural network with three layers of sizes 256, 128, and 1, and ReLU activation functions.
Using the manual features as the input, we trained the network with binary cross-entropy loss and Adam optimizer~\cite{kingma2014adam} for 12 epochs using the Keras library\footnote{\url{http://keras.io}}. We also experiment with the presence of dropout layers after the first two layers in order to prevent the model from overfitting the training data.


\paragraph{Hand-Crafted Features}

The following features are available for each tweet:
\begin{enumerate}[label={\alph*)}]
\item LexRank
\item Positive Sentiment
\item Negative Sentiment
\item Tweet Embedding
\end{enumerate}


Furthermore, a hashtag is represented by its own embedding.
For both the tweet and hashtag embeddings, we use 200-dimensional GloVe vectors, trained on 2 billion tweets\footnote{\url{http://nlp.stanford.edu/projects/glove/}}. Given the unique language of Twitter, we believe it is important to use Twitter-specific embeddings. The hashtag embedding is then the average of the individual hashtag tokens; the same holds true for the tweet embedding. If a token is not in the embedding corpus, its embedding defaults to the embedding for the `unknown' token.
For tweet tokenization we use a python wrapper for the ark-twokenizer\footnote{\url{https://github.com/myleott/ark-twokenize-py}}. We also believe that the use of embeddings trained on Twitter text will aid in providing external knowledge that is needed to perform at a high level in this task. For example, in Figure \ref{fig:tweets_example}, one tweet makes a reference to Harry Potter. Since an embedding for the token `potter' is present in the GloVe embeddings, this could potentially aid in the understanding of the tweet's humor.

\subsubsection{Recurrent Neural Network Model}
\label{sec:rnn}

Given the widespread effectiveness of recurrent neural networks for language modeling \cite{mikolov2010recurrent,sutskever2011generating,graves2013generating,bengio2006neural}, we implemented a token-level RNN-based model with the goal of learning better representations for both tweets and hashtags, which can be fed into the same FFNN as manual features. Given a sequence of tokens from either a tweet or a hashtag, we convert it into a sequence of GloVe vectors. Each sequence of vectors is fed into a Long Short-Term Memory unit (LSTM) \cite{hochreiter1997long}, which consumes an input vector at each time-step and produces a hidden state. The final hidden state constitutes the vector representation for the two tweets as well as the hashtag. We concatenate the three vector representations and provide it as input to the FFNN from the previous section. We apply rmsprop \cite{tieleman2012lecture} as the learning algorithm for this model.

\subsubsection{Character Level Model}
\label{sec:cnn}
Often, a joke in a tweet is based on a pun, such as combining two words to make a new `word'. For example, one of the top-10 tweets for the hashtag \textit{DogJobs} uses the word `barktender', combining the words `bark' and `bartender'. This property of the data leads to a high proportion of out of vocabulary (OOV) words. For example, in the GloVe embeddings that we used, the percentage of OOV tokens is 
32.27\%.
Since a token level model cannot understand single-token puns, we introduce a new character-level CNN model for this task
.

The model consists of two CNN layers of convolutions;  sized 5 and 3, each with 100 filters and max pooling of length 2. The input to the convolutions layers is a trainable character embeddings of size 50. The output of the CNN layers is passed to a fully connected layer of size 256. The representations of two tweets, learned by the these layers, are concatenated and fed to the same FFNN as in the previous section. 
\footnote{We also evaluated a character-level RNN model, which showed a similar performance while taking substantially longer to train.}

\subsubsection{Results}
\label{sec:results_supervised}

\begin{table*}[ht]
\centering
\scalebox{1}{
\begin{tabular}{|l|l|l|l|l|l|}
\hline
\textbf{System} & \textbf{Acc Micro Avg} & \textbf{Acc \textgreater 0.5} & \textbf{Max Acc} & \textbf{Min Acc} \\ \hline
Basic					& 0.513 ($\pm$0.0035)			& 53.5\% ($\pm$1.1628)			& 0.764 ($\pm$0.0185)			& 0.287 ($\pm$0.0187)			\\
Basic+HTE				& 0.501 ($\pm$0.0007)			& 48.4\% ($\pm$2.6854)			& 0.762 ($\pm$0.0165)			& 0.327 ($\pm$0.0084)			\\
Basic+TE				& 0.542 ($\pm$0.0007)			& 65.9\% ($\pm$0.6713)			& 0.769 ($\pm$0.0073)			& 0.310 ($\pm$0.0777)			\\
Basic+HTE+TE			& 0.546 ($\pm$0.0075)			& 69.8\% ($\pm$1.6444)			& 0.798 ($\pm$0.0294)			& 0.329 ($\pm$0.0068)			\\
Basic+HTE+RTE			& 0.502 ($\pm$0.0114) 			& 48.4\% ($\pm$10.422)			& 0.711 ($\pm$0.0265)			& 0.287 ($\pm$0.0257)			\\
Basic+HTE+TE+DRPT		& 0.554 ($\pm$0.0078)			& 72.1\% ($\pm$3.0765)			& 0.765 ($\pm$0.0361)			& 0.364 ($\pm$0.0327)			\\
HTE+TE					& 0.541 ($\pm$0.0058)			& 66.7\% ($\pm$3.5524)			& 0.749 ($\pm$0.0507)			& 0.371 ($\pm$0.0143)	\\ \hline
RNN (token-level)		& 0.554 ($\pm$0.0085)			& 73.3\% ($\pm$1.6444)			& 0.786 ($\pm$0.0779)			& 0.298 ($\pm$0.0150)			\\ \hline
CNN (character-level)	& \textbf{0.637} ($\pm$0.0074)	& 92.4 ($\pm$2.2076)	& \textbf{0.864} ($\pm$0.0515)	& 0.359 ($\pm$0.0401)			\\ \hline
RNN (character-level)	& 0.626 ($\pm$0.0017)	& \textbf{96.5\%} ($\pm$0.8772)	& 0.809 ($\pm$0.0134)	& \textbf{0.402} ($\pm$0.0318)			\\ \hline
\end{tabular}
} 
\caption{The results of the supervised experiments. Bold indicates the best system(s) according to the corresponding metric.}
\label{tbl:supervised_exp_results}
\end{table*}

%
%
%
%
%
%
%

The results of the supervised experiments are presented in Table~\ref{tbl:supervised_exp_results}. Because we assign labels to the training examples based on a random coin flip, we performed three runs for each system and present the average score (as well as the average for the other metrics). We also present the standard deviation of the three runs for a given system across the various metrics. The  feature types are as follows: Basic is the three features from the unsupervised experiments: lex rank, negative sentiment, and positive sentiment. HTE is an embedding for the hashtag from a specific file of tweets. TE are embeddings of the two tweets that constitute a single pair, one embedding for each tweet. DRPT indicates that we have added dropout of 0.5 between the fully connected layers of the FFNN. Because there is a noticeable performance gain when adding TE to the Basic+HTE system, this could potentially occur merely because of the added dimensions in the feature space. In order to address this, we experimented with the RTE feature, which is a random embedding for a given tweet, as opposed to the normal methodology for creating a tweet embedding. There are two types of RNN models: a token-level and a character-level model. Finally, CNN is the system described in Section~\ref{sec:cnn}. 
\section{Discussion}
The low accuracies of the unsupervised methodologies suggest that such a simple approach does not work for this complex task.
It is interesting to see that the positive sentiment and negative sentiment features perform almost identically. However, these features have quite strong negative correlation (-0.470). One hypothesis is that for certain hashtag, either positivity or negativity will play a more important role. The validity of this hypothesis is discussed further in relation to the supervised experiments below. Also of note in the results of the unsupervised experiments is the poor performance of LexRank. We believe this is because of the high variability of the language in the tweets, even within a specific hashtag. We also point out that in the work of \newcite{shahaf2015inside}, the authors report accuracy below 50\% when using n-gram perplexity as an unsupervised metric. The combination of these results dictate that language uniqueness is a poor unsupervised metric, regardless of dataset.

The complexity of this task, first revealed by the unsupervised experiments, is confirmed by the results of the supervised experiments. Two strong neural network models only surpassed random guessing by roughly 5\%.

One goal of the Basic system was to determine if a supervised system could learn an effective weighting of the three basic features, allowing it to outperform the results of the unsupervised experiments.
Another goal was to see if, by representing the hashtag, the system could learn for which hashtag a given basic feature is most important. However, the addition of the hashtag embedding to the Basic features actually creates a decrease in performance. One possibility is that the current hashtag representation is not able to facilitate the desired performance increase. Alternatively, the results show that the presence of the \textit{tweet} embedding creates a noticeable increase in performance. Having both together produces a very marginal increase in micro average, although the increase in percentage of hashtags with accuracy greater than 50\% is non-trivial. Furthermore, the poor performance of the system with the random tweet embedding shows that even averaging of individual token embeddings can provide a useful representation of a tweet's semantics.

The superior performance of the character-level models, compared to the performance of the token-level models, suggests that even a complex neural network system cannot perform well on this task using only token-level information.
Large amount of jokes in this dataset are based on puns, which leads to a large number of out of vocabulary words, even for embeddings trained on Twitter data. The fact that the character-level model performed substantially better than all other models suggests that this model can better represent OOV words (which, for example, is important for understanding puns) and use this information to decide which tweet is funnier.

While both systems recorded the same accuracy, it is interesting to note that the correlation of individual hashtag accuracies between the RNN and Basic+HTE+TE+DRPT systems is 0.557. This leads us to believe that even though the accuracies of the systems are the same, they are capturing different views of the data, and therefore perform better on different hashtags. This also suggests that an ensemble system could be effective for this task.

By comparing the performance of the RNN system with the HTE+TE system, we are able to see that in fact the RNN system is able to learn representations for the task that are more effective than simply averaging of token embeddings. We are able to make this claim by the fact that the representations learned by the RNN system feed into the same FFNN as the feature-based approach.


One final analysis we perform is to determine if the test hashtags that are most similar to the training hashtags actually perform better than those that are less similar. To determine this, we represent a hashtag by its average embedding. We then hold a given hashtag out and calculate the cosine similarity with the average of the remaining hashtags' embeddings. This represents how similar a test hashtag is to the remaining hashtags for training. We then calculate the correlation between this similarity and the accuracy of the test hashtag. We did this for the results of the Basic+HTE+TE+DRPT system. Unfortunately, the correlation is relatively low (0.223). However, this low correlation could also be explained by the fact that averaging of individual tokens for the hashtag doesn't appear to be the appropriate representation for this task.

Lastly, we note the stability of results for the same systems across multiple runs. None of the systems (aside from the one with the random tweet embedding) have a standard deviation in micro accuracy above 0.01, which shows that even by randomly assigning labels to the dataset, the better systems are able to distinguish themselves.

\section{Conclusion}
In this paper, we have presented the HW humor dataset. We motivate the need for such a dataset, while also describing our collection process. Our dataset is several orders of magnitude greater than the only existing comparable dataset, the NYCC dataset. Lastly, we present the results of both supervised and unsupervised experiments. 
The results of our experiments show that this task cannot be solved with a simple token-level approach, and requires a more complex system working with puns at the character level in order to solve the task with an accuracy  that is substantially greater than random guessing.

There are numerous avenues for future work. 
We acknowledge that responding to these hashtags often requires external knowledge, such as titles of movies or names of bands. Our results show that semantic representations alone cannot capture this. In such cases, this external knowledge is mandatory to understanding why a tweet is funny. Systems that make effective use of external knowledge sources will have a better chance to recognize the humor in a tweet and will therefore have higher performance in this task. 

An ambitious implementation for interacting with external knowledge sources is a Neural Turing Machine (NMT) \cite{graves2014neural}. Interacting with a knowledge source requires discrete actions, such as querying/not querying, as well as deciding on the query string. \newcite{zaremba2015reinforcement} describe an algorithm for training an NTM with discrete interfaces. For example, an NTM might learn, for a given hashtag, which specific external knowledge source would be beneficial for deciphering the humor in response tweets, as well as how to determine which part of a tweet string refernces which external knowledge. Consequently, our dataset is of secondary interest for researchers who seek to interact with query interfaces via NTMs.

\bibliography{tacl}
\bibliographystyle{acl2012}

\end{document}